\documentclass[runningheads]{llncs}
\usepackage{graphicx}
\usepackage{hyperref}%
\usepackage{multirow}%
\usepackage{amsmath,amssymb,amsfonts}%
\usepackage{mathrsfs}%
\usepackage[title]{appendix}%
\usepackage{xcolor}%
\usepackage{textcomp}%
\usepackage{manyfoot}%
\usepackage{booktabs}%
\usepackage{algorithm}%
\usepackage{algorithmicx}%
\usepackage{algpseudocode}%
\usepackage{listings}%
\usepackage{multicol}%
\usepackage{subcaption}%
\usepackage{lipsum,lineno}%

\begin{document}
\title{S+t-SNE - Bringing Dimensionality Reduction to Data Streams}%\thanks{Supported by organization x.}}
%
%\titlerunning{Abbreviated paper title}
% If the paper title is too long for the running head, you can set
% an abbreviated paper title here
%
\author{Pedro C. Vieira\inst{1,4,5}\orcidID{0000-0002-0546-0124} \and
João P. Montrezol\inst{1,4}\orcidID{0000-0003-1346-216X} \and
João T. Vieira\inst{1}\orcidID{0009-0006-3206-2910}
\and João Gama\inst{2,3}\orcidID{0000-0003-3357-1195}}
\authorrunning{P. C. Vieira et al.}
% First names are abbreviated in the running head.
% If there are more than two authors, 'et al.' is used.
%
\institute{Department of Computer Science, Faculty of Sciences, University of Porto
\and
Faculty of Economics, University of Porto \\
\and
INESC TEC \\
\email{\{pedrocvieira, joao.antunes, up201905419\}@fc.up.pt, jgama@fep.up.pt}
\and
Equal contribution, order defined by coin flip.
\and
Corresponding author
}
\maketitle              % typeset the header of the contribution
\begin{abstract}
We present S+t-SNE, an adaptation of the t-SNE algorithm designed to handle infinite data streams.
The core idea behind S+t-SNE is to update the t-SNE embedding incrementally as new data arrives, ensuring scalability and adaptability to handle streaming scenarios. By selecting the most important points at each step, the algorithm ensures scalability while keeping informative visualisations. By employing a blind method for drift management, the algorithm adjusts the embedding space, which facilitates the visualisation of evolving data dynamics. Our experimental evaluations demonstrate the effectiveness and efficiency of S+t-SNE, whilst highlighting its ability to capture patterns in a streaming scenario. We hope our approach offers researchers and practitioners a real-time tool for understanding and interpreting high-dimensional data.

\end{abstract}

%%%%%%%%%%%%%%%%%%%%%%%%%%%%%%%%%%%%%%%%%%%%%%%%%%%%%%%%%%%%%%%%%%%%%%%%%%%%%%%%%%%%%%
\section{Introduction}\label{sec:intro}
Dimensionality reduction techniques are an object of great interest in applications such as image or natural language processing.
Dimensionality reduction techniques simplify complex data, ensuring better interpretability of such data and helping its visualisation. Furthermore, they enhance model performance by employing feature selection and thus improving computation speed
Constructing efficient algorithms for dimensionality reduction in a streaming context opens the possibility of working with potentially infinite datasets. Hence, such algorithms could be used with arbitrary-size datasets, offline or online. For example, it would be possible to visualise static, very large datasets typically used in deep learning. Another example application would be to use this algorithm to improve computation speed and provide a human-readable visualisation of data streams in services like electricity, gas, and water maintenance.

In this paper, we explain the limitations of existing dimensionality reduction techniques when applied to data streams and propose a new method named S+t-SNE to solve some limitations.

In section \ref{sec:Stsne}, we explain how our approach works and handles common challenges, such as the continuously increasing volume of historic data and the constant flow of new data entries, together with the possibility of concept change. In section \ref{sec:exp}, we delve into the tests performed to evaluate the performance of the proposed algorithm. The version used for the tests is available in a code repository\footnote{\href{https://github.com/PedrV/S--t-SNE}{github.com/PedrV/S--t-SNE}} and follows implementation specifics to be integrated with the River\footnote{\href{https://riverml.xyz}{riverml.xyz}} framework, as described in the aforementioned framework's documentation.
%%%%%%%%%%%%%%%%%%%%%%%%%%%%%%%%%%%%%%%%%%%%%%%%%%%%%%%%%%%%%%%%%%%%%%%%%%%%%%%%%%%%%%

%%%%%%%%%%%%%%%%%%%%%%%%%%%%%%%%%%%%%%%%%%%%%%%%%%%%%%%%%%%%%%%%%%%%%%%%%%%%%%%%%%%%%%
\section{Related Work}\label{sec:sota}

The t-distributed stochastic neighbour embedding (t-SNE) \cite{Maaten2008} specialises in transforming a high-dimensional dataset into a two or three-dimensional dataset. 
It does so by using a t-distribution as a basis for calculating the similarity between points in the projected space and the original space while using the KL divergence to guide the point positioning. 

Dimensionality reduction techniques are classified into ``out-of-sample'' and ``in-sample'' categories \cite{bengio2003out}. Out-of-sample techniques start with a small data subset and map the rest accordingly, making them more scalable but less accurate when the subset does not accurately represent the full dataset. In-sample techniques, like t-SNE, classical Multidimensional Scaling (MDS), and UMAP \cite{umap}, process the entire dataset at once, resulting in more accurate results but with higher computational costs. The algorithm discussed in this paper falls into the out-of-sample category.

Developments in out-of-sample techniques focus on incorporating user knowledge into the projection process. Examples include Piecewise-Laplacian Projection (PLP) \cite{paulovich2011piece}, Least Squares Projection (LSP) \cite{paulovich2008least}, and Local Affine Multidimensional Projection (LAMP) \cite{joia2011local}. While these techniques can handle larger datasets, they are often unsuitable for data streams, as they rely on the quality of the initial data subset. Some online strategies have been developed to mitigate this problem. Basalaj \cite{basalaj1999incremental} introduces an online version of classical Multidimensional Scaling. In his work, when a new data entry is received, MDS is applied considering both the new entry and the already processed ones to create a new full pairwise distance matrix. Alsakran et al. \cite{alsakran2011streamit} apply a force-based approach, updating to consider new instances and recomputing the full pairwise distance matrix in memory. Jenkins et al. \cite{jenkins2004spatio} and Law et al. \cite{law2004nonlinear} introduce online versions of the ISOMAP in-sample technique. 
Upon receiving a new data entry, every previous entry is also processed, and a full pairwise distance matrix must be computed. Law et al. \cite{law2006incremental} make this process faster by sidestepping the requirement of a full pairwise matrix in their introduction of an online version of LMDS. In this technique revision, the only distances the algorithm needs to compute are the ones between the new data entry and the pre-existing entries. Kouropteva et al. \cite{kouropteva2005incremental} and Schuon et al. \cite{schuon2008truly} present online revisions of the in-sample LLE technique through the definition of strategies that update neighbourhood relationships when a new entry is introduced. 
Rauber et al. \cite{rauber2016visualizing} introduce a Dynamic t-SNE. This technique facilitates the projection of windows of datasets that depend on time while maintaining spatial consistency in the positions of points in projections. Although this has interesting results, Dynamic t-SNE must re-project \textit{all} received data to create an entirely new projection to compute the most recent data. Consequently, it is necessary to keep the whole dataset in memory, which is inadequate for streaming scenarios in which the data continues to grow. This is something that the method proposed in this paper attempts to address. Furthermore, there is, to the best of our knowledge as of writing, no attempt made by any of the methods described to deal with concept drift. Our approach will address this problem.
%%%%%%%%%%%%%%%%%%%%%%%%%%%%%%%%%%%%%%%%%%%%%%%%%%%%%%%%%%%%%%%%%%%%%%%%%%%%%%%%%%%%%%

%%%%%%%%%%%%%%%%%%%%%%%%%%%%%%%%%%%%%%%%%%%%%%%%%%%%%%%%%%%%%%%%%%%%%%%%%%%%%%%%%%%%%%
\section{Streaming t-SNE (S+t-SNE)}\label{sec:Stsne}

In a streaming scenario, data arrives continuously. This assumption hinders the application of the traditional t-SNE. The two inherent challenges to this application are (1) the duration or termination point of the data stream is often unknown, rendering it impossible to determine when to halt the accumulation of points and start the application of t-SNE; (2) the potential accumulation of points due to an extensive or even infinite data stream poses obstacles on the computation time and the computational resources. 

\subsection{Problem One - When to start}\label{subsec:p1}
One possible approach to using t-SNE in a streaming scenario involves accumulating all encountered points until a change in the data stream is detected, at which point the accumulated points are projected. 
However, for this technique to work, the data would need to exhibit drift and we would have to establish a threshold of ``how much drift is enough drift''.
An alternative is to adopt a fixed batch-wise approach to mitigate these challenges in our work. Points are accumulated until a predetermined batch size, $\mathcal{B}$, is attained. Subsequently, t-SNE is applied to project the accumulated points. This approach offers a swift and agnostic solution that does not rely on specific data patterns, enabling its off-the-shelf application.

One iteration of S+t-SNE consists of accumulating a new data point, checking if the total accumulated is equal to $\mathcal{B}$, and if so, applying t-SNE. The first projection is made by applying t-SNE to the batch of data. Since there are no points in the projection space, normal t-SNE will be applied in the first projection. However, after the first iteration of S+t-SNE, subsequent iterations will project in a space where points already exist.

In our approach, incorporating new data points from iteration $t+1$ into the projection space involves considering conditional probabilities between new and previously embedded points from iteration $t$. 
To achieve the intended outcome, our approach is grounded in the openTSNE framework \cite{Poliar2019}, with a primary reliance on the technique of partial embedding. The concept of partial embedding facilitates the incorporation of new points into an established embedding space by considering only the conditional probabilities between points in $t$ and $t+1$.
However, one limitation of this approach is that focusing solely on these conditional probabilities may omit the natural inclusion of conditional probabilities between new data points. Hence, groups with similar conditional probabilities to already embedded points yet exhibiting low conditional probabilities between themselves may converge into the same area of the lower-dimensional space. This constraint arises from relying solely on information from old points for new point embeddings. The method from subsection \ref{subsec:drift} will mitigate this effect by removing unnecessary points.

\subsection{Problem Two - Reduction of space fingerprint}\label{subsec:p2}
Section \ref{subsec:p1}, overcomes the issue of determining when to apply t-SNE in the streaming context (Problem 1). However, the concern of accumulating points in the projected space persists. 

To address the accumulation of points in the lower-dimensional space, we propose retaining the shape of groups of points by using a clustering algorithm applied to t-SNE projections. Each group's shape will be represented by convex hulls (the calculation of a convex hull is done in $O(n \, log \, n)$, where $n$ is the size of a cluster), minimising the number of retained points. However, convex hull points are not informative enough for new point incorporations. We introduce ``\textit{PEDRUL}'' (Points Expected to Define Regions of Unambiguous Location) to represent important points. 

The PEDRUL within each 2D group is determined by their density in the original D-dimensional space. Hence, for each group of points in the embedding space, their PEDRUL is defined as the points in the original space with the largest number of neighbours within a search radius, denoted as $R$. To control the transitivity of the neighbourhood of dense points, the search identifies points as candidates for being PEDRUL only if they are not in a neighbourhood of an already defined PEDRUL point.
This methodology may sacrifice the selection of \textit{de facto} densest points but maintains the integrity and unambiguity of the dense regions that each PEDRUL spans within each group.

To efficiently obtain the PEDRUL, the KDTree \cite{DBLP:journals/corr/cs-CG-9901013} data structure is used for nearest neighbour searches in multi-dimensional spaces. KDTree has a construction time complexity of $O(n \; log \, n)$ and a balanced structure (the height of the tree does not exceed $O(log \, n)$). 

Once the KDTree structure is constructed, it can be queried to retrieve all points within a specified distance from a given point. 
Each point in the current batch's neighbourhood limited on $R$ is estimated and sorted by size. The resulting neighbours are stored in a hashable set, enabling efficient intersection calculations and allowing the identification and selection of sparse density points as required.

In summary, PEDRUL reduces the number of points in the low-dimensional space by preserving only points with maximal information (maximal points around them) that will be used for subsequent embeddings. To aid visualisation and limit regions, we store the points delimiting the convex hull of each group. Accumulating too many PEDRUL will not be a problem since for an entire application of our algorithm, the number of PEDRUL remains constant. If the initial indication is to hold 100 points, there will be 100 points in the projection at all times. However, the points can change across iterations.

\subsection{Handling Drift}\label{subsec:drift}

Data streams often have non-stationary distributions, with drift taking the form of sudden or gradual changes. Sudden drift is an abrupt shift in data distribution without temporal overlap between the pre and post-change distributions. Gradual drift involves a slower, incremental overlapping transition \cite{Agrahari2022}. Adapting t-SNE for online use requires addressing these drift types. 
In the following paragraphs, we propose a method that can be coupled with incremental t-SNE proposed earlier and with any method using convex regions. The proposed method is fit for online scenarios and handles drift by updating the projections in the space of interest - the low-dimensional space. Furthermore, it mitigates the possible artefacts referred to in section \ref{subsec:p2}.
When sudden drift occurs, data embeddings are likely to experience steep changes. Hence, the old embeddings correspond to old views of the data and must be immediately removed. Gradual drift necessitates gradually removing examples as they become less relevant. Adjusting for different drifts is connected with the concept of forgetting different points at different speeds.

\begin{figure}[!t]
    \centering
    \includegraphics[width=0.5\textwidth]{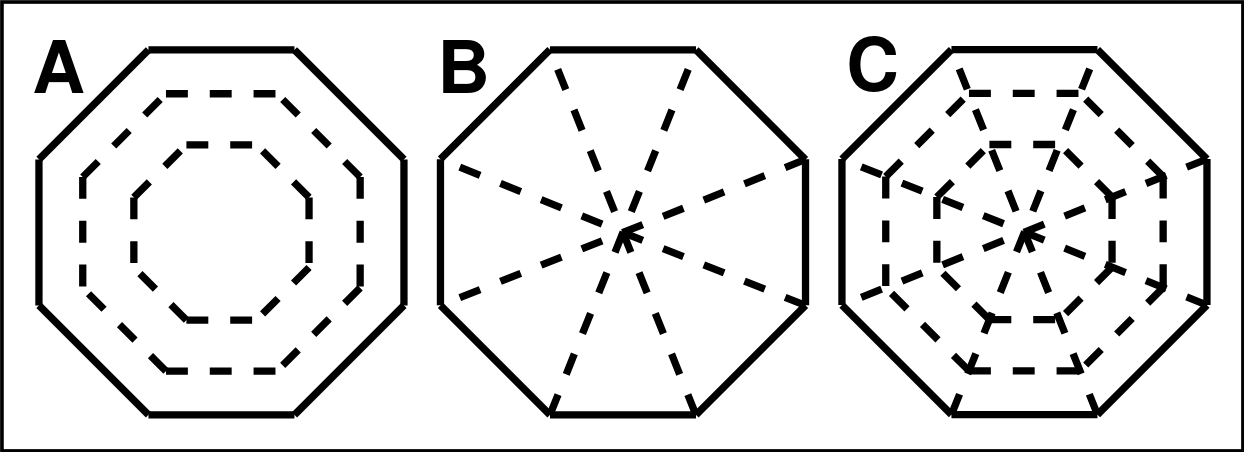}
    \caption{In shape A) we see the concentric cuts over a convex hull; In B) the median cuts; In C) all cuts together, forming a structure similar to a cobweb.}
    \label{fig:ecs_cuts}
\end{figure}

Our solution involves using the convex hulls obtained from clustering and dividing them into parts. Each partition will employ blind drift detection by exponential decay based on the number of iterations in S+t-SNE. This allows parts without new points during a period (given by exponential decay) to disappear, ensuring consistency. 
We parameterise the exponential decay with three parameters, $\alpha = 0.88, \beta = 1.6$ and $\eta = 0.01$, yielding $N(t) = \alpha e^{-t\eta + \beta}$ where $t$ is the number of iterations. This expression encapsulates our definition of drift. In this configuration, a polygon in iteration 200 will have section $x$ cut if said section does not receive points for more than $N(200)$ iterations.
The selected partitions are along the medians of the polygon (figure \ref{fig:ecs_cuts} - B) and its concentric regions (figure \ref{fig:ecs_cuts} - A). These partitions allow us to monitor any translation with arbitrary precision as long as enough iterations are completed. Furthermore, they allow for arbitrary deletion while ensuring the result is always a convex shape. Since shapes are always convex, maintaining the algorithm is efficient. 

Let $n$ denote the maximum number of vertices of a polygon from the embedding space, $k$ the number of polygons, $p$ the maximum number of PEDRUL registered in a polygon, and $m$ the number of concentric regions.
The temporal complexity to determine the median regions is $\Theta(n)$, testing ownership of a point to a region is $\Theta(1)$. Performing a cut takes $O(n)$, where $n$ is the number of vertices of the polygon. A cut is done by deleting points counter-clockwise from one end of a point of a median region to another. Maintaining all median regions of all polygons takes $O(2\, p\, k\, n)$. An important fact to speed up this procedure is that all operations can be efficiently parallelisable polygon-wise. This means that all polygons can run in parallel, reducing the total time to independent of the number of polygons and linear on the number of PEDRUL.
Calculating a concentric region for a polygon has a $\Theta(n)$ complexity.
As for testing for point membership, it takes $O(log \, n)$ using a Delaunay tessellation of triangles. Constructing such a tessellation takes $O(n \; log \, n)$, but this is only done once per region.
The total iteration complexity, including the construction of the tesselation, takes $O(m \; k \; p \; log \, n) + O(m \; k \;  n \; log \,n)$. The factor $m$ is generally small ($\leq 5$). Hence, we will disregard it in asymptotic analysis. Like the median regions, the concentric regions have highly independent operations, making them highly parallelisable. A parallel implementation reduces the complexity to $O(p \; log \, n) + O(n \; log \,n)$, which is effectively $O(p \; log \, n)$. 
From this point onwards, we will refer to the method described above as Exponential Cobweb Slicing (ECS).
The interplay between S+t-SNE and ECS is displayed in algorithm \ref{alg:itsne-ecs}.
Points that suffer from the problem described in \ref{subsec:p1} are projected onto the wrong region $A$ of the space. If new points that truly belong to $A$ appear, the algorithm self-corrects. If no points appear mapped to $A$ then ECS will trigger and cut $A$ out. Hence, the visual artefacts are reduced.

\begin{algorithm}[!t]
\caption{S+t-SNE}\label{alg:itsne-ecs}
\begin{algorithmic}[1]
\Require \textit{conn} is alive 
\Ensure $\text{dataStorage} = \emptyset$

\State $\text{proj} \gets$ \Call{EmptyProjection}{NULL}

\While{True}
\State newPoint $\gets$ \Call{ReceiveData}{$\text{conn}$} 
\State dataStorage $\gets$ \Call{StoreData}{newPoint}
\If{dataStorage.length $= \mathcal{B}$}
    \State proj $\gets$ \Call{ProjectPoints}{proj, dataStorage}
    % \Statex
    \State proj.PEDRUL $\gets$ \Call{CalculatePEDRUL}{proj} \Comment{For new projections}
    \State proj.hulls $\gets$ \Call{CalculateConvexHulls}{proj} \Comment{For visualization}
    \State dataStorage $\gets \emptyset$
    % \Statex
    \State proj.iterations $\gets$ proj.iterations + 1 
    \State proj.hulls $\gets$ \Call{ECS}{proj.hulls, proj.iterations}
    \For{PEDRUL \textbf{in} proj.PEDRUL}
        \If{PEDRUL \textbf{not in} proj.hulls}
            \State \Call{RemovePoint}{proj.PEDRUL, PEDRUL} 
        \EndIf
    \EndFor
\EndIf
\EndWhile
\end{algorithmic}
\end{algorithm}

%%%
%%%%%%%%%%%%%%%%%%%%%%%%%%%%%%%%%%%%%%%%%%%%%%%%%%%%%%%%%%%%%%%%%%%%%%%%%%%%%%%%%%%%%%

%%%%%%%%%%%%%%%%%%%%%%%%%%%%%%%%%%%%%%%%%%%%%%%%%%%%%%%%%%%%%%%%%%%%%%%%%%%%%%%%%%%%%%
\section{Experiments}\label{sec:exp}
Our methodology does not compare itself with the alternatives delineated in section \ref{sec:sota} due to those lacking an available implementation by the authors, blocking, from our point of view, a fair comparison between algorithms.
However, we evaluate and compare S+t-SNE against t-SNE. With these tests, we aim to understand the strengths and weaknesses of the proposed method in comparison to its original variant. Hopefully, the comparison against a strong baseline demonstrates the strength of S+t-SNE.
All tests used a Windows 10 Pro 22H2 system, with an AMD Ryzen 7 5800x 3.8 GHz (boosting to 4.5GHz) processor and 32Gb (3200 MHz) RAM.

\paragraph{Datasets.} We use two datasets: MNIST, used in \cite{kouropteva2005incremental,law2006incremental,rauber2016visualizing}, and a synthetic dataset to evaluate drift. 
The latter dataset was created by randomly selecting points from three different spaces (structures) of dynamic 3D distributions. Each distribution is configured with spatial movement, adjusting both mean and covariance parameters at each time step emulated by a tick count
and consist of 525000 points or 175000 per structure. Two of the structures used will suffer translation, contractions, and dilations in space, and the other will only experience the last two effects. The structures overlap in the high-dimensional space, increasing the difficulty of the dataset. The purpose of this dataset is to assess the performance of S+t-SNE in a scenario closer to a real online one.

\paragraph{Configurations.} All datasets will be streamed, emulating an online paradigm for data acquisition. 
The parameters for t-SNE were used empirically, resulting in the best-looking projections.
For t-SNE to work, we adopt batch projections akin to those utilized in S+t-SNE. Specifically, upon the accumulation of $\mathbb{B}$ data points, t-SNE reprojects the entire dataset. Consequently, the last reprojected iteration encapsulates a conventional application of t-SNE to the entire dataset, similar to an offline approach. 
The parameters for the internal t-SNE used in S+t-SNE were the same as the ones used in the pure t-SNE. 
Said parameters can be viewed in the code repository. The number of iterations for t-SNE is 500 rounds of optimisation and 250 of early exaggeration. 
These settings allow for the overall best results. 
Regarding the number of iterations, factoring the time for ECS, S+t-SNE using 400 optimisation iterations and 250 early exaggeration iterations is the comparable setting to t-SNE. 
Furthermore, the parameters regarding the ECS and the selection of density points can be revised in the repository. 
Since using different parameters yields cuts at different iterations for the same dataset, hence a different ``definition'' of drift, we opted to keep the parameters from section \eqref{subsec:drift} as they present good results visually.

\subsection{Results}
\begin{figure}[!t]
    \centering
    \includegraphics[width=0.9\textwidth]{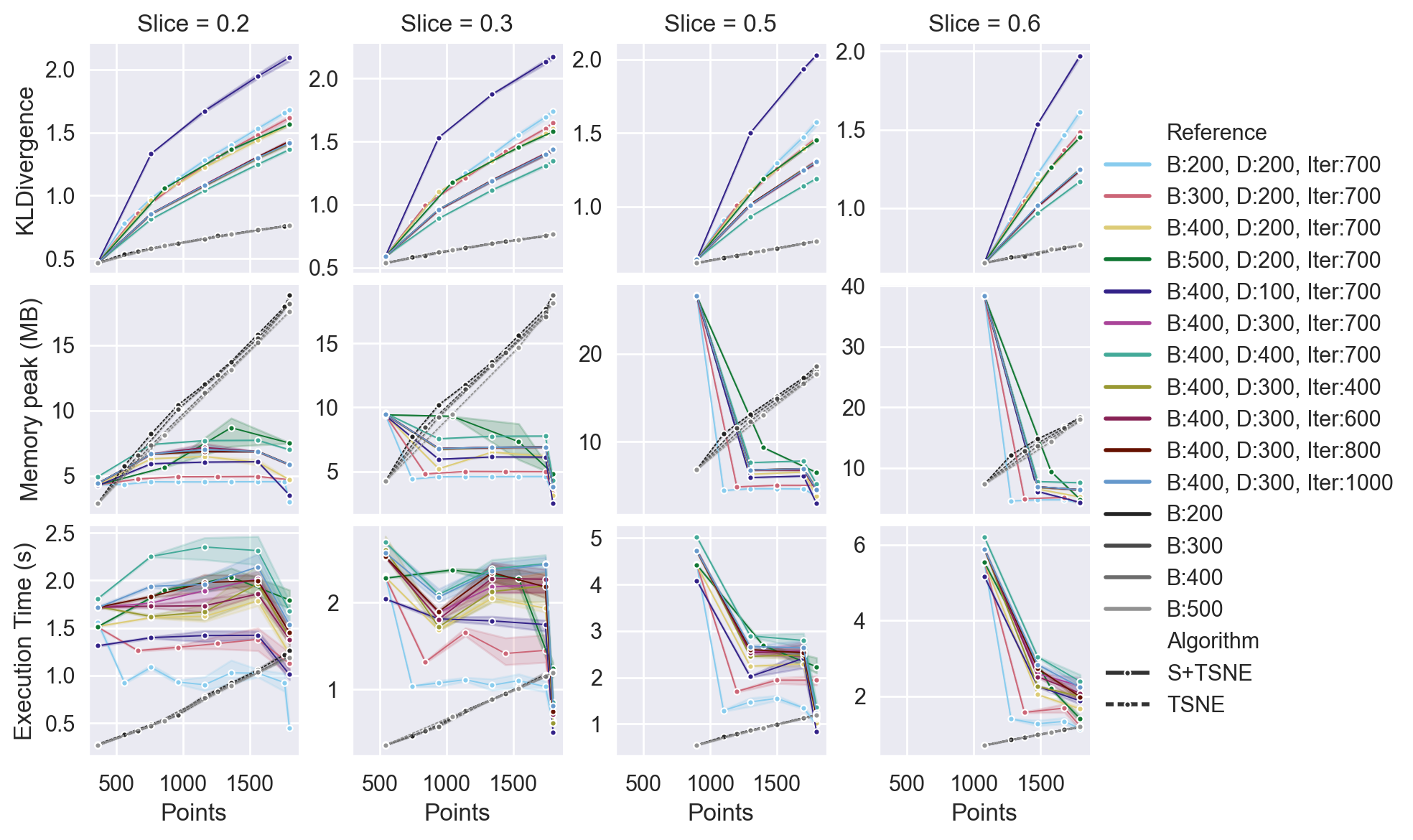}
    \caption{Break-down of KLD, Peak Memory and Time for the MNIST dataset.}
    \label{fig:mnist_analysis}
\end{figure}

\paragraph{MNIST.} Figure \ref{fig:mnist_analysis} shows the curves for comparing the algorithms on the MNIST dataset. The continuous lines and the t-SNE by the dashed lines represent the S+t-SNE algorithm. Each point in a line represents an action by the algorithm, typically at the end of the batch. The main point of comparison is the initial slice of the data used to define the opening space. That is the quantity of data accumulated before projecting the first time. The slice is represented fractionally regarding the total amount of data from $0$ to $1$. The lines representing S+t-SNE represent the variation in the size of batch (B), the number of PEDRUL (or density points) considered (D),
and the number of iterations allowed the algorithm to run (Iter). 

\paragraph{KLD.} We used the Kullback–Leibler divergence (KLD) to measure the entropy between the projection and the points in the original space. Looking at figure \ref{fig:mnist_analysis}, we see that the KLD for t-SNE increases as the number of points increases, meaning that having more points deteriorates the t-SNE performance. This result is not unexpected since more points can cause more entropy. As for S+t-SNE, all configurations present a slow increase in the divergence as the number of points increases. Based on our analyses, the global rate of change manifests, in the least favourable scenario, as a progressively slow function relative to the number of batches processed by the S+t-SNE algorithm (figures in the repository). We see that larger batches and more PEDRUL cause a smaller KLD. A larger number of PEDRUL has more influence than larger batches. This may happen because more PEDRUL points mean more anchors for incoming points to have as a reference, meaning better projections. Technically, t-SNE uses all points of the dataset as PEDRUL, serving as a benchmark for performance.

\paragraph{Memory.} As expected, the peak memory used by t-SNE (figure \ref{fig:mnist_analysis}) aggressively increases as the number of points increases, achieving the same memory peak in all slices in the last reprojection. As for S+t-SNE, the peak memory use is achieved when getting the opening projection. The initial S+t-SNE iteration (getting the opening projections) for slice $x$ uses the same process of a t-SNE application for the same slice $x$. However, the peak memory of S+t-SNE is higher because it has to account for the search of the PEDRUL points. We notice that larger slices further increase the initial memory peak gap between algorithms because searching for PEDRUL points within more data is more costly. After the first iteration, the peak memory reduces drastically and remains constant.

\paragraph{Time.} The results for execution time (figure \ref{fig:mnist_analysis}) were as expected. Larger batches, number of points and iterations increase the computational time. Furthermore, a large slice increases the initial time of S+t-SNE but contributes to a faster decrease in the time consumed.

\begin{figure}[!t]
    \centering
    \includegraphics[width=1\textwidth]{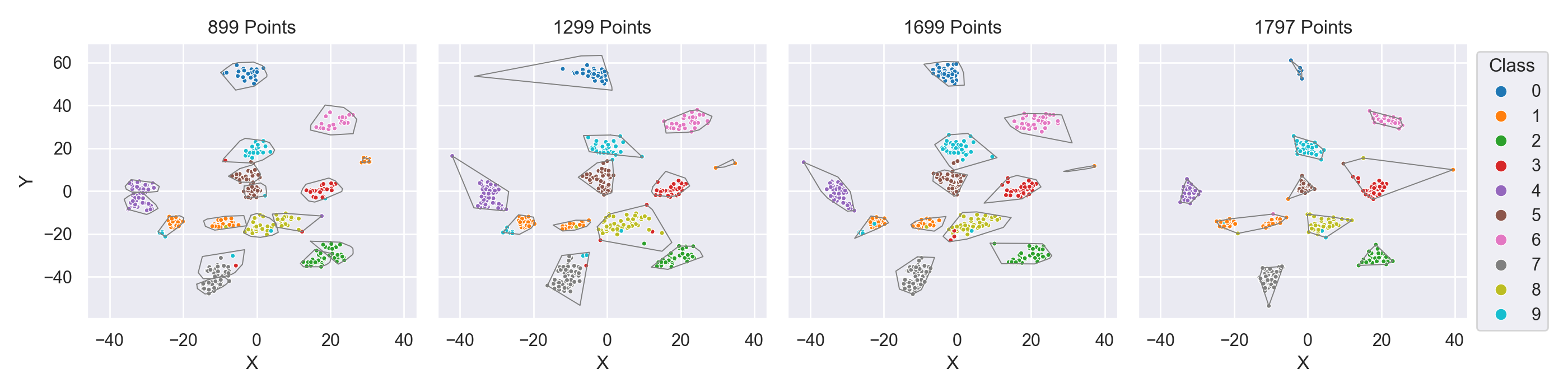}
    \caption{Projections by S+t-SNE for MNIST. B:400 D:400 IT:700 Slice:0.2}
    \label{fig:mnist_hulls}
\end{figure}

Knowing the final drop in memory and time for S+t-SNE is due to reaching the end of the dataset, for the general case, based on this dataset, we would want to have the initial slice as large as possible. We should hold as many points as possible before the initial projection, increasing the quality of the representation by trying to obtain a stable enough opening projection. However, we must consider the larger memory footprint and initial computation time for larger slices.
The number of PEDRULs and batches should be as large as possible until the limit of memory and time is available. However, more PEDRUL is highly preferred to a larger batch. We believe the number of iterations has a negligible influence. Hence, we advocate for selecting the minimum iteration count necessary for convergence.

\begin{figure}[!t]
    \centering
    \includegraphics[width=0.9\textwidth]{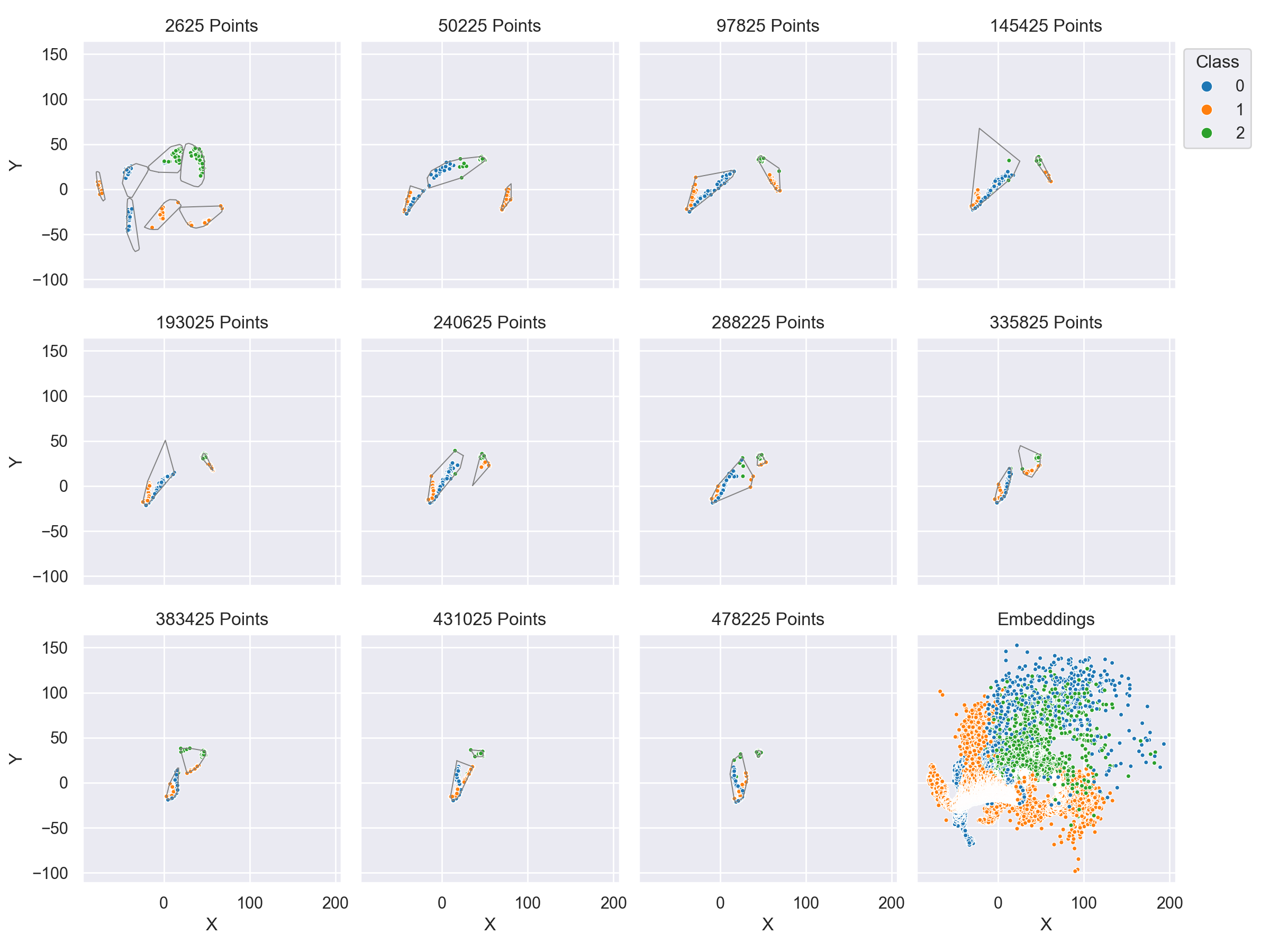}
    \caption{S+t-SNE for synthetic data and total accumulated points (bottom-right).}
    \label{fig:drifted_analysis}
\end{figure}

Figure \ref{fig:mnist_hulls} shows the evolution of the S+t-SNE embeddings for MNIST for B:400 D:400 IT:700 Slice:0.2. Besides the points delimiting the hulls, the PEDRUL are also visible. All ten classes are well distinguishable. Moreover, ECS eliminates inaccurately projected points, as posited in section \ref{subsec:p1}. One example is the transition from the top-right figure to the bottom-left, where the blue cluster had its stretched side removed. It's a typical ECS cut.
\paragraph{Synthetic Data.} Figure \ref{fig:drifted_analysis} shows the results obtained for the synthetic dataset. We were not able to run t-SNE past the initial slice. Hence, we ran S+t-SNE only with a slice size of 0.005. Furthermore, it was impossible to calculate the KLD due to the size of the matrix.

\begin{figure}[!t]
    \centering
    \includegraphics[width=0.9\textwidth]{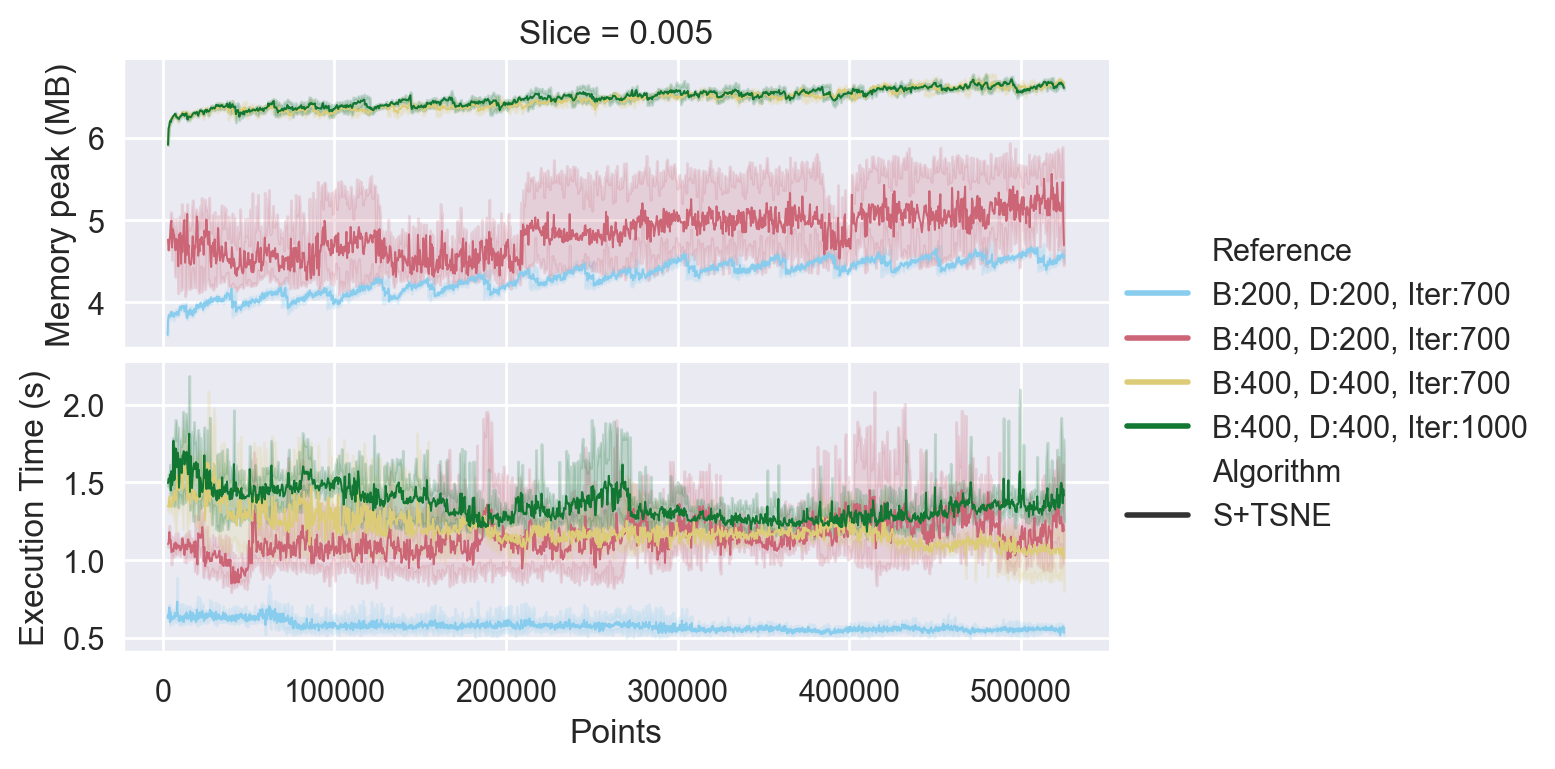}
    \caption{Break-down of KLD, Peak Memory and Time for the dataset with drift.}
    \label{fig:drifted_analysis-stats}
\end{figure}

\paragraph{Memory.} From figure \ref{fig:drifted_analysis-stats}, we can see that increasing the batch size and number of points increases the peak memory consumption. However, the increase is very small (in the order of single bytes), even though we are doubling batch size and number of points. This result happens %possibly
due to the capacity of larger batches and the larger number of points, allowing us to pick points with better representational capabilities, lasting more iterations and triggering fewer operations and consequently less accumulation of objects in memory. 
The results confirm the ones obtained with MNIST: the memory used is near constant after reaching a stable point where the representation is stable (initial peak removed in image).
\paragraph{Time.} Figure \ref{fig:drifted_analysis-stats} corroborates the findings derived from the MNIST dataset. Notably, the quasi-constant computational time after the first iteration (initial peak removed in image).

We believe the considerations taken from MNIST regarding the value to use for each parameter hold in this case. Figure \ref{fig:drifted_analysis} shows the projections for the synthetic data. The bottom-right plot shows the embeddings if no ECS was used and all points were considered as PEDRUL. The plots show that our techniques allow us to focus on the points of interest. Hence, we can see the most recent movements of the dataset, these being the gradual shrinking of the green structure and the orange and blue ones slowly drifting closer to each other until they overlap.

The effect of ECS is not noticeable in the computational time due to its small 
runtime footprint. Even without a parallel implementation, the main difference between configurations is the number of hulls/clusters found at each iteration. However, this number is similar to all configurations because the clustering parameters and dataset are the same for all configurations. Furthermore, the number of PEDRUL has a constant effect on the time taken by ECS, and since the gap between configurations in this regard is not very large, the effect is negligible. The drift has virtually no impact on memory consumption because of its gradual nature. 
%%%%%%%%%%%%%%%%%%%%%%%%%%%%%%%%%%%%%%%%%%%%%%%%%%%%%%%%%%%%%%%%%%%%%%%%%%%%%%%%%%%%%%

%%%%%%%%%%%%%%%%%%%%%%%%%%%%%%%%%%%%%%%%%%%%%%%%%%%%%%%%%%%%%%%%%%%%%%%%%%%%%%%%%%%%%%
\section{Conclusion}\label{sec:conc}
In this paper, we developed an efficient adaptation of t-SNE called S+t-SNE to work with data streams. Our version supports dimensionality reduction of online data and can adapt to data drift. 
A possible direction for future work is to test different mechanisms for obtaining PEDRUL and test the robustness of the method to different types of drift. To allow for a uniform comparison across online methods, it would be interesting to develop a metric for the comparison of projections of arbitrary size.
%%%%%%%%%%%%%%%%%%%%%%%%%%%%%%%%%%%%%%%%%%%%%%%%%%%%%%%%%%%%%%%%%%%%%%%%%%%%%%%%%%%%%%

%
%
%

%
% ---- Bibliography ----
%
% BibTeX users should specify bibliography style 'splncs04'.
% References will then be sorted and formatted in the correct style.
%
\noindent
{\bf Acknowledgements} We thank the APPIA-Portuguese Association for Artificial Intelligence for financial support. Jo\~ao Gama acknowledges the support of the project AI-BOOST, funded by the European Union under GA No 101135737.

\bibliographystyle{splncs04}

\begin{thebibliography}{10}
\providecommand{\url}[1]{\texttt{#1}}
\providecommand{\urlprefix}{URL }
\providecommand{\doi}[1]{https://doi.org/#1}

\bibitem{Maaten2008}
{Visualizing Data using t-SNE}. Journal of Machine Learning Research  \textbf{9}(86),  2579--2605 (2008)

\bibitem{Agrahari2022}
Agrahari, S., Singh, A.K.: Concept drift detection in data stream mining : A literature review. Journal of King Saud University - Computer and Information Sciences  \textbf{34}(10),  9523--9540 (Nov 2022)

\bibitem{alsakran2011streamit}
Alsakran, J., Chen, Y., Zhao, Y., Yang, J., Luo, D.: Streamit: Dynamic visualization and interactive exploration of text streams. In: 2011 IEEE Pacific Visualization Symposium. pp. 131--138. IEEE (2011)

\bibitem{basalaj1999incremental}
Basalaj, W.: Incremental multidimensional scaling method for database visualization. In: Visual Data Exploration and Analysis VI. vol.~3643, pp. 149--158. SPIE (1999)

\bibitem{bengio2003out}
Bengio, Y., Paiement, J.f., Vincent, P., Delalleau, O., Roux, N., Ouimet, M.: Out-of-sample extensions for lle, isomap, mds, eigenmaps, and spectral clustering. Advances in neural information processing systems  \textbf{16} (2003)

\bibitem{jenkins2004spatio}
Jenkins, O.C., Matari{\'c}, M.J.: A spatio-temporal extension to isomap nonlinear dimension reduction. In: Proceedings of the twenty-first international conference on Machine learning. p.~56 (2004)

\bibitem{joia2011local}
Joia, P., Coimbra, D., Cuminato, J.A., Paulovich, F.V., Nonato, L.G.: Local affine multidimensional projection. IEEE Transactions on Visualization and Computer Graphics  \textbf{17}(12),  2563--2571 (2011)

\bibitem{kouropteva2005incremental}
Kouropteva, O., Okun, O., Pietik{\"a}inen, M.: Incremental locally linear embedding. Pattern recognition  \textbf{38}(10),  1764--1767 (2005)

\bibitem{law2006incremental}
Law, M.H., Jain, A.K.: Incremental nonlinear dimensionality reduction by manifold learning. IEEE transactions on pattern analysis and machine intelligence  \textbf{28}(3),  377--391 (2006)

\bibitem{law2004nonlinear}
Law, M.H., Zhang, N., Jain, A.K.: Nonlinear manifold learning for data stream. In: Proceedings of the 2004 SIAM International Conference on Data Mining. pp. 33--44. SIAM (2004)

\bibitem{DBLP:journals/corr/cs-CG-9901013}
Maneewongvatana, S., Mount, D.M.: Analysis of approximate nearest neighbor searching with clustered point sets. CoRR  \textbf{cs.CG/9901013} (1999)

\bibitem{umap}
McInnes, L., Healy, J., Melville, J.: Umap: Uniform manifold approximation and projection for dimension reduction (2018)

\bibitem{paulovich2008least}
Paulovich, F.V., Nonato, L.G., Minghim, R., Levkowitz, H.: Least square projection: A fast high-precision multidimensional projection technique and its application to document mapping. IEEE Transactions on Visualization and Computer Graphics  \textbf{14}(3),  564--575 (2008)

\bibitem{paulovich2011piece}
Paulovich, F.V., Eler, D.M., Poco, J., Botha, C.P., Minghim, R., Nonato, L.G.: Piece wise laplacian-based projection for interactive data exploration and organization. In: Computer Graphics Forum. vol.~30, pp. 1091--1100. Wiley Online Library (2011)

\bibitem{Poliar2019}
Poli{\v{c}}ar, P.G., Stra{\v{z}}ar, M., Zupan, B.: {openTSNE}: a modular python library for t-{SNE} dimensionality reduction and embedding  (Aug 2019)

\bibitem{rauber2016visualizing}
Rauber, P.E., Falcao, A.X., Telea, A.C., et~al.: Visualizing time-dependent data using dynamic t-sne.  (2016)

\bibitem{schuon2008truly}
Schuon, S., Durkovi{\'c}, M., Diepold, K., Scheuerle, J., Markward, S.: Truly incremental locally linear embedding. In: CoTeSys 1st International Workshop on Cognition for Technical Systems (2008)

\end{thebibliography}

\end{document}